\documentclass[sigconf]{acmart}
\AtBeginDocument{%
  }

\setcopyright{acmlicensed}
\copyrightyear{2025}
\acmYear{2025}
\acmDOI{XXXXXXX.XXXXXXX}
\acmConference[CHIWORK '25]{Symposium on Human-Computer Interaction for Work}{June 23--25,
  2025}{Amsterdam, Netherlands}
\acmISBN{978-1-4503-XXXX-X/18/06}

\usepackage{soul}
\usepackage{subcaption}

\begin{document}

\title{Humble AI in the real-world: the case of algorithmic hiring}

\author{Rahul Nair}
\email{rahul.nair@ie.ibm.com}
\orcid{0000-0002-1875-2409}
\author{Inge Vejsbjerg}
\email{ingevejs@ie.ibm.com}
\orcid{0000-0003-3039-2140}
\author{Elizabeth Daly}
\email{elizabeth.daly@ie.ibm.com}
\orcid{0000-0003-0162-3683}
\affiliation{%
  \institution{IBM Research}
  \city{Dublin}
  \country{Ireland}
}
\author{Christos Varytimidis}
\email{varytimidis@workable.com}
\orcid{0009-0004-8025-9051}
    \affiliation{
    \institution{Workable}
    \city{Athens}
    \country{Greece}
    }
\author{Bran Knowles}
\email{b.h.knowles1@lancaster.ac.uk}
\orcid{0000-0002-2554-1896}
\affiliation{
    \institution{Lancaster University}
    \city{Lancaster}
    \country{United Kingdom}
}

\renewcommand{\shortauthors}{Nair et al.}

\begin{abstract}
    Humble AI (Knowles et al., 2023)
    argues for cautiousness  in AI development and deployments through scepticism (accounting for limitations of statistical learning), curiosity (accounting for unexpected outcomes), and commitment (accounting for multifaceted values beyond performance). We present a real-world case study for humble AI in the domain of algorithmic hiring. Specifically, we evaluate virtual screening algorithms in a widely used hiring platform that matches candidates to job openings. There are several challenges in misrecognition and stereotyping in such contexts that are difficult to assess through standard fairness and trust frameworks; e.g., someone with a non-traditional background is less likely to rank highly. We demonstrate technical feasibility of how humble AI principles can be translated to practice through uncertainty quantification of ranks, entropy estimates, and a user experience that highlights algorithmic unknowns. We describe preliminary discussions with focus groups made up of recruiters. Future user studies seek to evaluate whether the higher cognitive load of a humble AI system fosters a climate of trust in its outcomes.
\end{abstract}



\maketitle

\section{Introduction}
\label{sec:intro}

Our first interaction with work is through the hiring process. Firms solicit applications  from candidates for a job opening by putting out an advertisement. Recruiters then screen applications for suitability for the role. Shortlisted candidates are then interviewed and a final selection made. While specific details may vary from firm to firm, these are generally the multiple steps involved in the hiring pipeline. 

Increasingly, firms are turning to algorithmic tools for these tasks. Dedicated human resource (HR) platforms are used for managing job requisitions, candidate pools, r\'{e}sum\'{e} parsing, and a wide range of activities that would typically be conducted by human recruiters. There are over 250 AI tools for various HR functions \cite{wef2021}. This proliferation of automation tools brings with it serious challenges related to bias. Labour markets are rife with structural inequities \cite{bertrand2004emily} related to race, gender, age and other candidate characteristics. There is growing recognition that AI automation can amplify and scale up these biases (see \cite{fabris2024fairness} for a recent survey on fairness challenges in hiring). The prime example of such cases in practice is the machine learning recruiting tool used by Amazon that discriminated against women \cite{reuters2018}. 

Even without algorithms, making good hiring decisions is a difficult proposition. ``Hiring is about probabilities'' \cite{Sink2006} where information signals about candidates serve as proxies for how well they are likely to do in the future. These signals can be misleading. Candidates that are good on paper can turn out to be poor fit for the role, and conversely those with less impressive credentials turn out to be stellar.

Taken together, the increasing use of AI automation to assist with tasks that are subjective and difficult for humans within the hiring process brings a risk of amplifying bias, and motivates reconsideration about how such systems ought to be designed and developed. \textit{Humble AI} principles were designed to foster trust in AI by reducing harm to individuals who are passed over for opportunities due to an inability of the model to recognize their eligibility.

\begin{figure*}[ht]
    \centering
    \captionsetup[subfigure]{justification=centering}
    \begin{subfigure}[t]{0.45\textwidth}
            \centering
            \includegraphics[width=\linewidth,height=50mm, keepaspectratio]{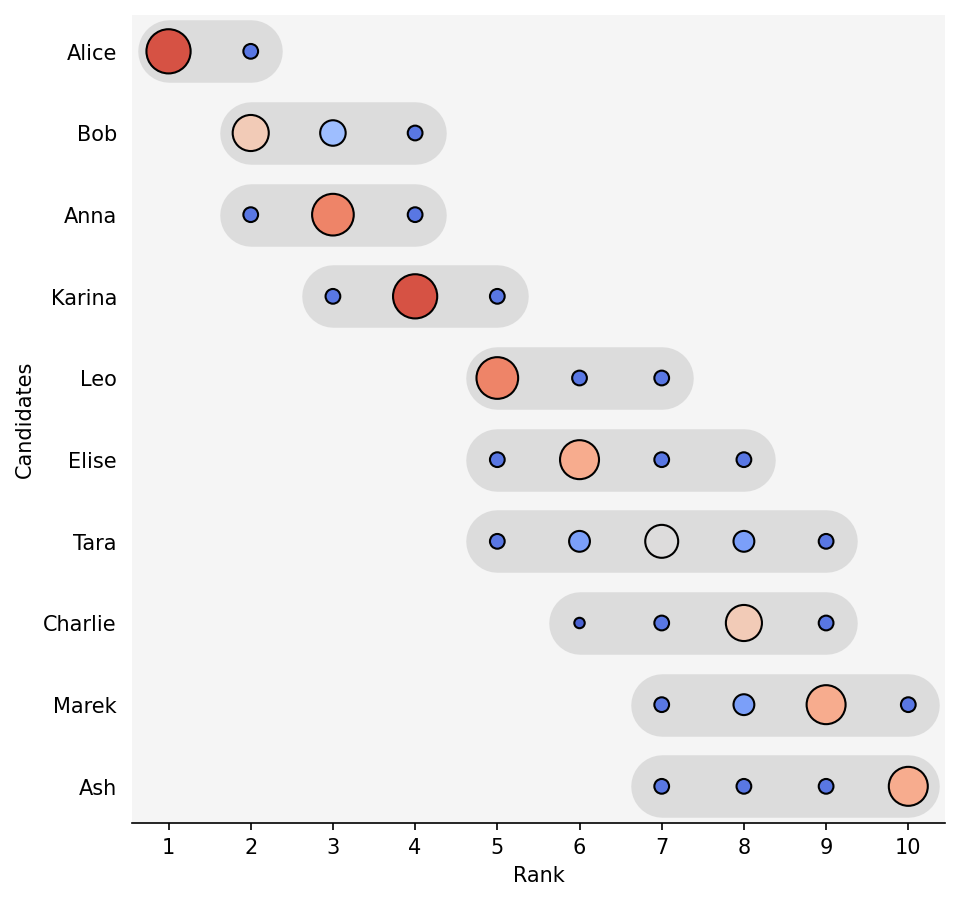}
            \caption{Low uncertainty regime where AI system \\is less uncertain on candidate rankings}
            \label{fig:intro:ranksets:lo}
        \end{subfigure}
    \begin{subfigure}[t]{0.45\textwidth}
            \centering
            \includegraphics[width=\linewidth,height=50mm, keepaspectratio]{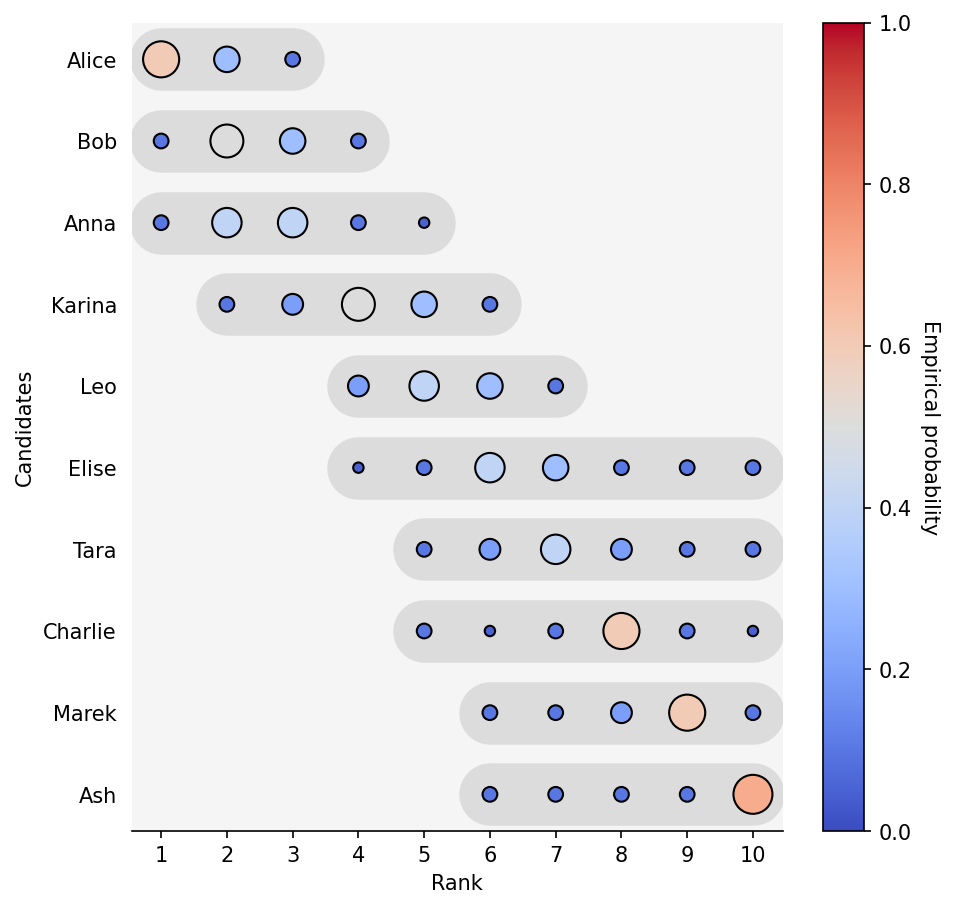}
            \caption{High uncertainty regime where AI system\\has high uncertainty}
            \label{fig:intro:ranksets:hi}
        \end{subfigure}
    \caption{Rank sets for two prototypical scenarios}
    \label{fig:intro:ranksets}
    \Description{Image describing the concept of ranksets for two cases. Left shows rank sets with low uncertainty, and right image shows rank sets with high uncertainty.}
\end{figure*}

In this work, we explore the principles of \emph{Humble AI} \cite{knowles2023humble} in this hiring context and see how these can be applied in practical settings. Humble AI argues for cautiousness in AI deployments through three principles. The first is \emph{scepticism} in model predictions as they are inherently uncertain and being ``open to the possibility of being wrong''. The second principle is \emph{curiosity} by exploring what-if scenarios, if alternative decisions to those made by an AI system were considered. The last principle is \emph{commitment} to a value system that goes beyond model performance and considers broader processes that foster trust. 

Specifically, we focus on one step of the hiring pipeline, candidate screening. Given a job description and a (large) pool of candidates, the screening step seeks to determine a ranking of the candidates based on suitability for that specific position. This is typically accomplished using a recommendation system or a machine learning model that predicts scores for each candidate that indicate hiring potential. The top ranked candidate is adjudged to be the best suited for the job. The top-$k$ candidates are moved on further through the hiring pipeline.

We operationalize \emph{Humble AI} principles with uncertainty and entropy quantification procedures and design a user experience that exposes AI uncertainty and confidence to end users. Since screening is effectively a ranking problem, our primary innovation here is around the use of \emph{rank sets}, a generalisation of ranking candidates. Instead of a fixed ranking, a rank set denotes a set of possible ranks a candidate can take along with an associated probability for each rank. Figure \ref{fig:intro:ranksets} shows prototypical examples of when AI system estimates of candidate rankings are less or more uncertain. Considering probability of ranking allows us to compute uncertainty of an AI system for specific candidates. An additional measure of entropy (see the Section \ref{sec:methods} for more details) further quantifies candidates for which the AI system's decisions are less reliable. Candidates with high rank entropy may need additional manual reviews.

We demonstrate key ideas on a widely used real-world HR platform\footnote{\texttt{<anonymised>}} that performs virtual screening. Using actual job specifications and anonymised candidate profiles. Additionally, we report on synthetic data experiments where the true ranking is known. In the presence of noise, we demonstrate that accounting for uncertainty improves the rate at which the true rankings can be recovered. Preliminary feedback from a focus group of recruiters and HR specialists highlight gaps in our designs and challenges in the user experience.

As contributions, we operationalise principles of \textit{Humble AI} in a practical real-world setting. Using the idea of rank sets, we demonstrate how uncertainty and entropy measures capture key elements of model ignorance. We propose methods to compute rank sets in black box settings where AI system access is only available for inference, and lastly we report on preliminary focus group discussions with recruiters on our envisioned user experience.

\section{Methods}
\label{sec:methods}

Given a set of $n$ candidates $\mathcal{X} = \{ x_1, x_2, \ldots, x_n \}$ for a specific job $y$ and a scoring function $f_y: \mathcal{X} \rightarrow \mathbb{R}$, the basic screening task seeks a permutation $\sigma$ that orders the $n$ candidates based on the scores in descending order, i.e. $f_y(x_{\sigma(1)}) \geq f_y(x_{\sigma(2)}) \geq \dots \geq f_y(x_{\sigma(n)})$. We assume that higher scores are more desirable and indicate greater relevance of a candidate for a job. Additionally, the scoring function is assumed to be available in a black-box manner. In other words, the AI system can be only used for inference.

In normal cases, the virtual screening involves presenting the recruiter with the top-$k$ candidates from the permutation $\sigma$. In our humble proposal, we seek an alternative permutation $\sigma_h$ that accounts for uncertainty in the scoring function. To achieve this, instead of considering $z_i=f_y(x_i)$ as being the point estimate of the score, we consider $\tilde{z}_i$ to be a random variable with a known empirical distribution. To estimate this distribution, we perform local perturbations for each candidate and evaluate the scoring function for each perturbation. Our perturbation mechanism samples in the neighbourhood of the candidate vector $x_i$ and is similar to the explainability method LIME \cite{lime}. Specifically, perturbations are done by random feature masking. This gives us a set of empirical score distributions for each candidate, $\{\tilde{z}_1, \tilde{z}_2, \ldots, \tilde{z}_n\}$.

A rank set is defined by a probability matrix $P = [p_{ij}]$, where $p_{ij}$ denotes the probability of the $i$-th candidate taking on the $j$-th rank. This idea is inspired by work done in ranking language models \cite{chatzi2024prediction}. While in their case, they define it to be a bounded range of ranks, we relax this constraint.  

To estimate $p_{ij}$, we run a monte-carlo simulation by making independent draws. Each draw provides a permutation of ranks, aggregating across all draws gives us the rank probabilities. This procedure can result in elements of $P$ having very low probabilities. One can ignore low values using an appropriate threshold (0.01 in our experiments). 

\begin{figure*}[ht]
    \centering
    \captionsetup[subfigure]{justification=centering}
    \begin{subfigure}[t]{0.45\textwidth}
            \centering
            \includegraphics[width=\linewidth]{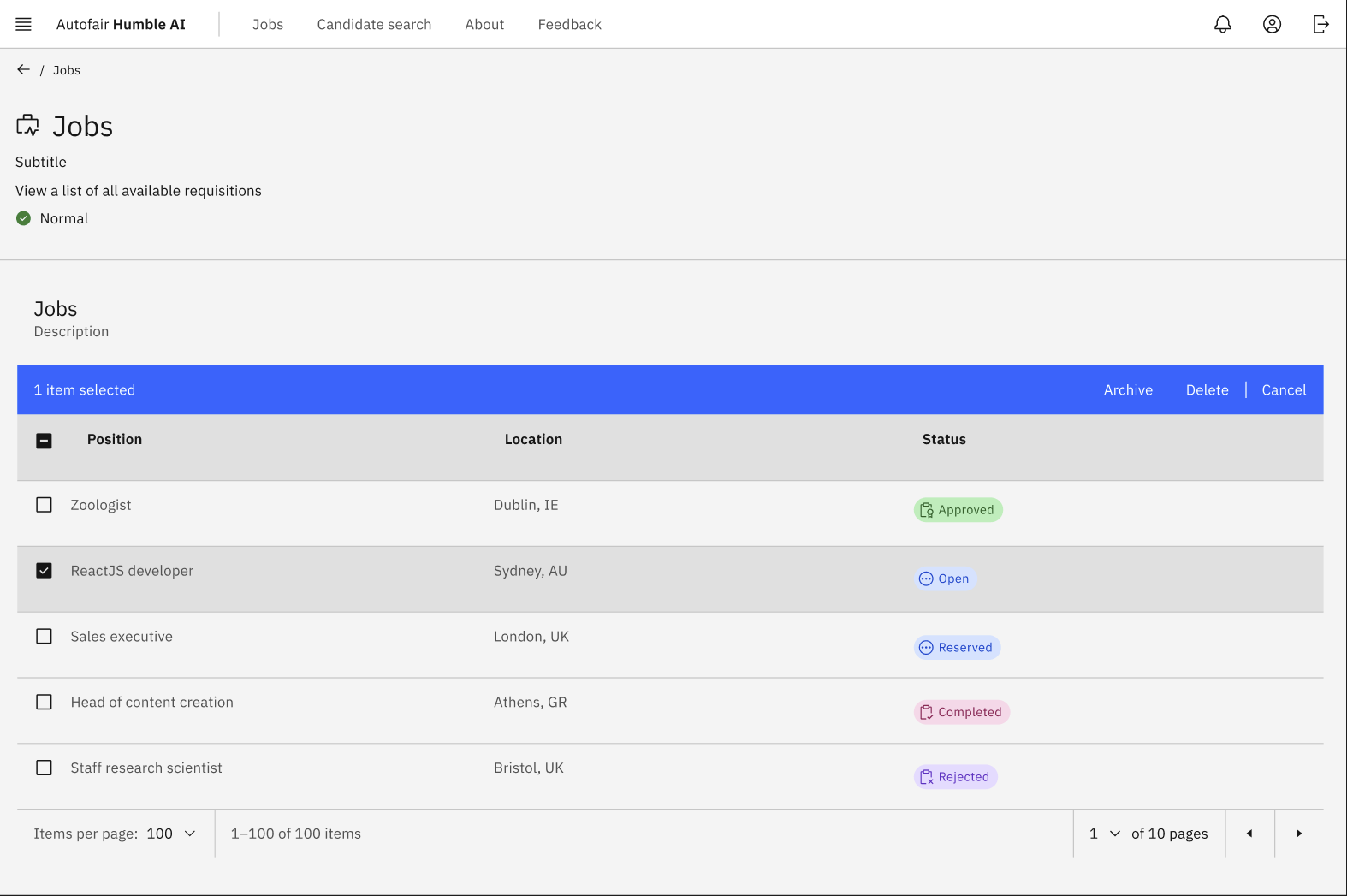}
            \caption{Job listings}
            \label{fig:methods:ux:joblisting}
        \end{subfigure}
    \begin{subfigure}[t]{0.45\textwidth}
            \centering
            \includegraphics[width=\linewidth]{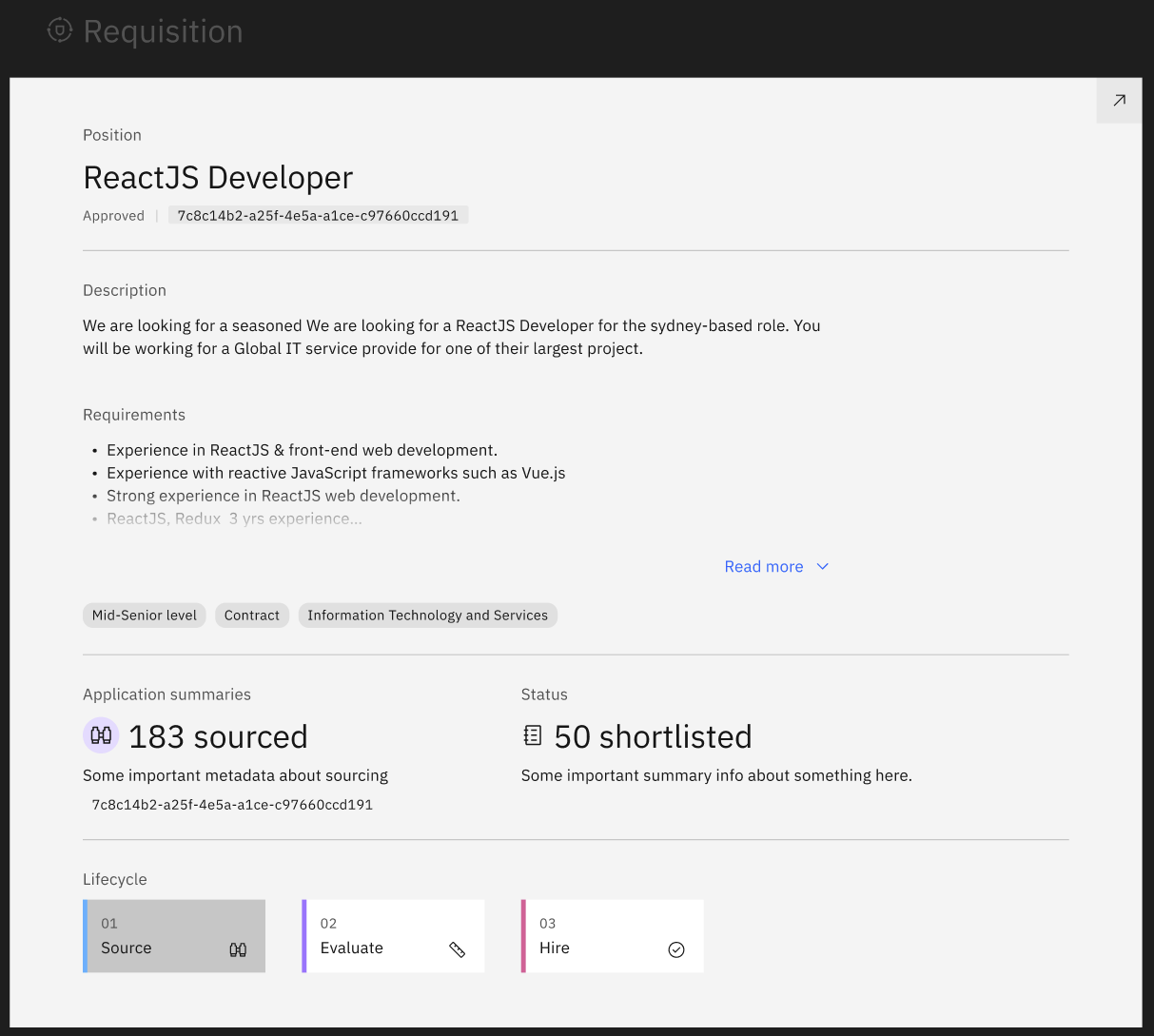}
            \caption{Job details}
            \label{fig:methods:ux:jobdetails}
        \end{subfigure}
    \caption{User experience for job specifications}
    \label{fig:methods:ux:job}
    \Description{Image describing user experience in defining a job requisition and details on job advertisement.}
\end{figure*}

\begin{figure*}[ht]
    \centering
    \captionsetup[subfigure]{justification=centering}
    \begin{subfigure}[t]{0.45\textwidth}
            \includegraphics[width=\linewidth]{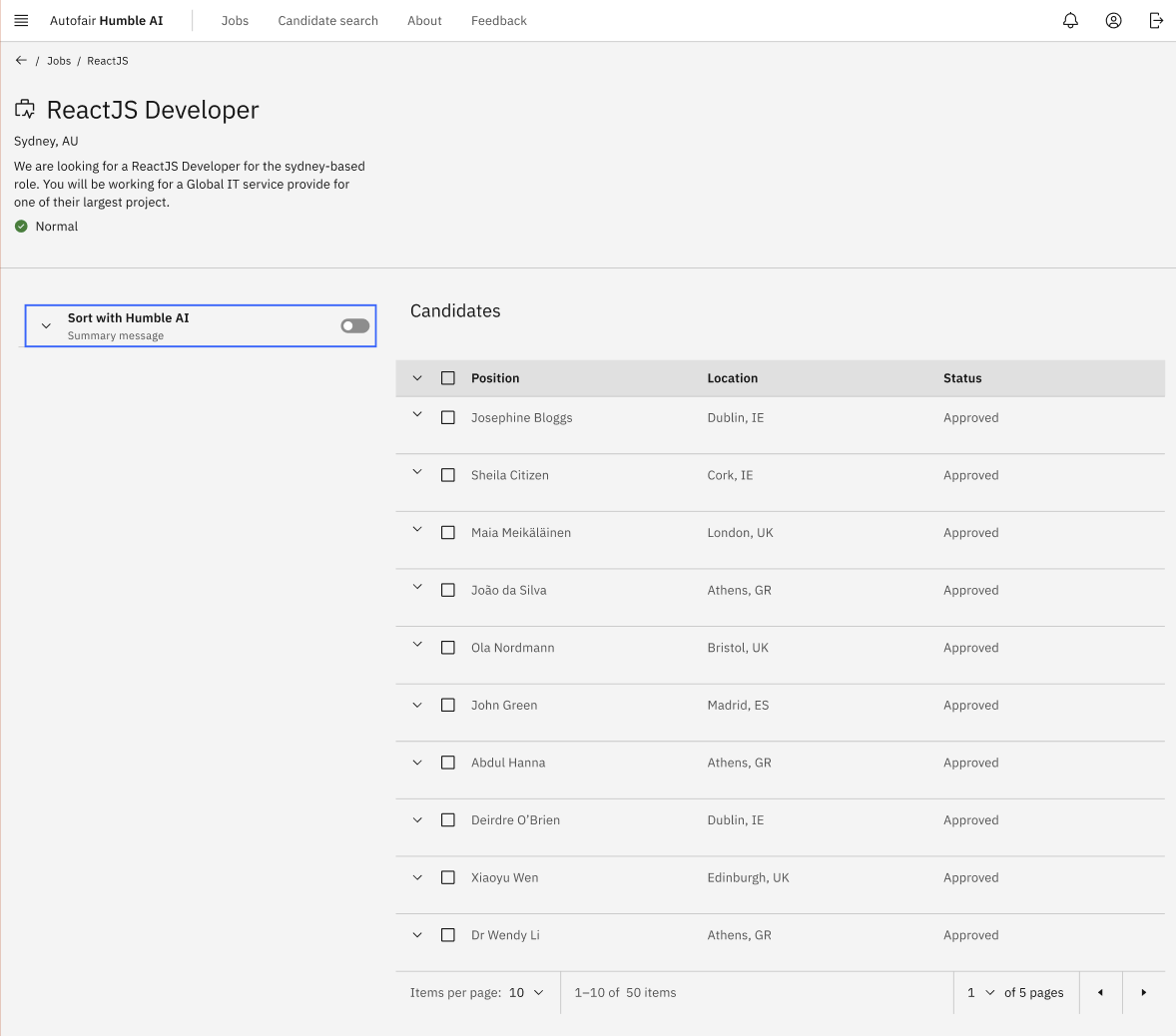}
            \caption{Candidate matches (deterministic)}
            \label{fig:methods:ux:candidates:deterministic}
        \end{subfigure}
    \begin{subfigure}[t]{0.45\textwidth}
            \includegraphics[width=\linewidth]{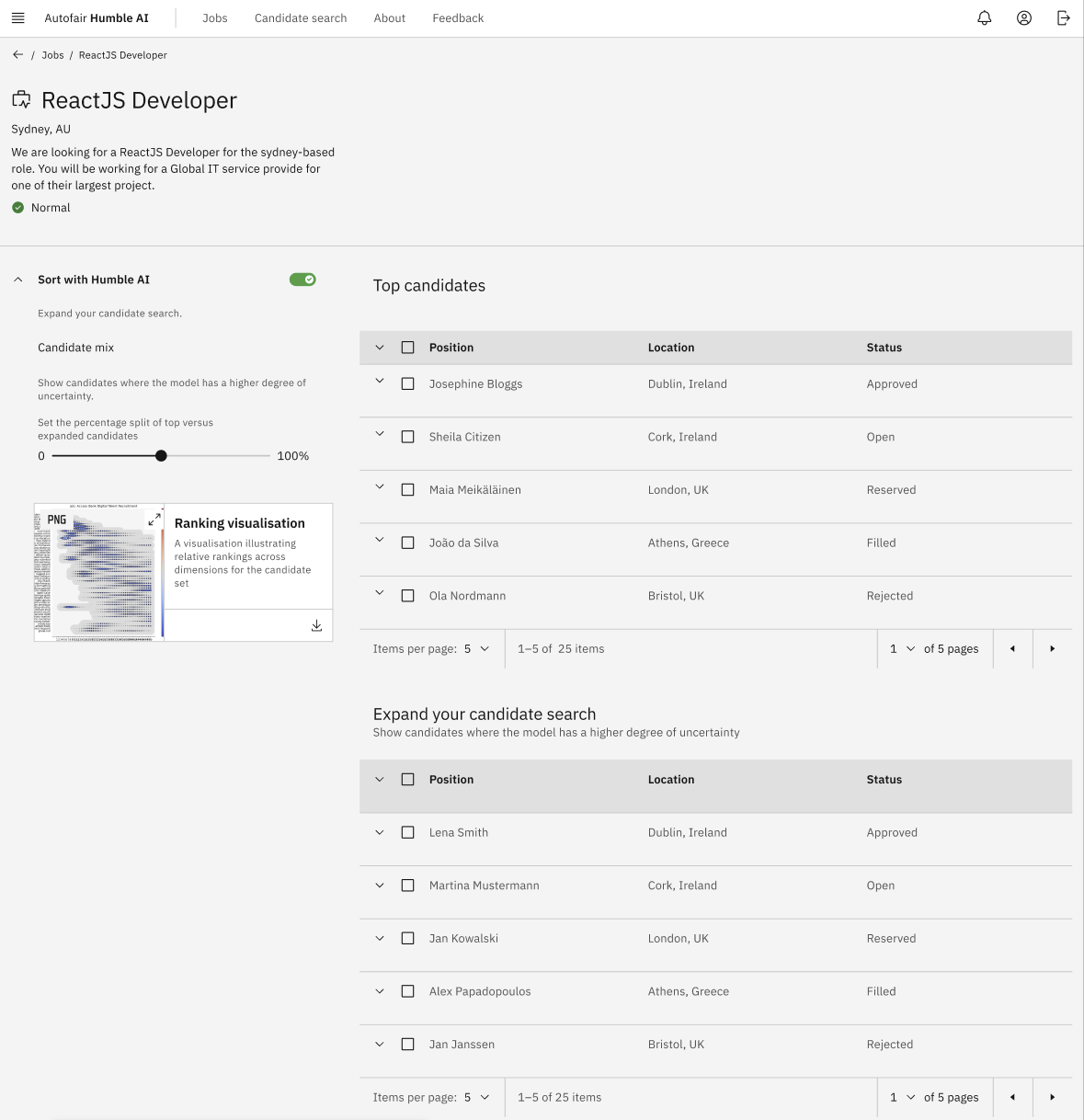}
            \caption{Candidate matches with humble AI principles}
            \label{fig:methods:ux:candidates:uncertain}
        \end{subfigure}
    \caption{User experience for candidate screening}
    \label{fig:methods:ux:candidates}
    \Description{Image describing user experience in candidate screening.}
\end{figure*}

Three quantities are of interest are computed: (a) the expected rank of the $i$-th candidate, i.e. $\sum_{j \in n} p_{ij} j$, (b) the entropy for each candidate $i$, i.e. $\sum_{i\in n} p_{ij} \log p_{ij}$, and (c) rank variance. We additionally experimented with determining the most likely ranking based on linear programming. However this turned out to be uninformative when rank variance is high.

\begin{figure*}[ht]
    \centering
    \captionsetup[subfigure]{justification=centering}
    \begin{subfigure}[t]{0.48\textwidth}
            \centering
            \includegraphics[width=\linewidth,height=50mm, keepaspectratio]{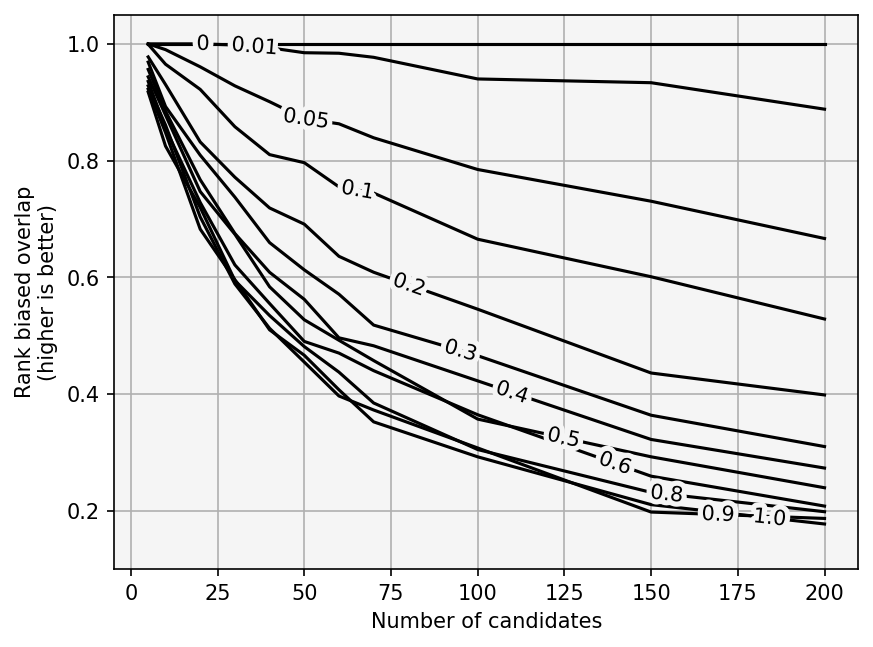}
            \caption{Performance of ranking based on point estimates}
            \label{fig:methods:synth:deterministic}
        \end{subfigure}
    \begin{subfigure}[t]{0.48\textwidth}
            \centering
            \includegraphics[width=\linewidth,height=50mm, keepaspectratio]{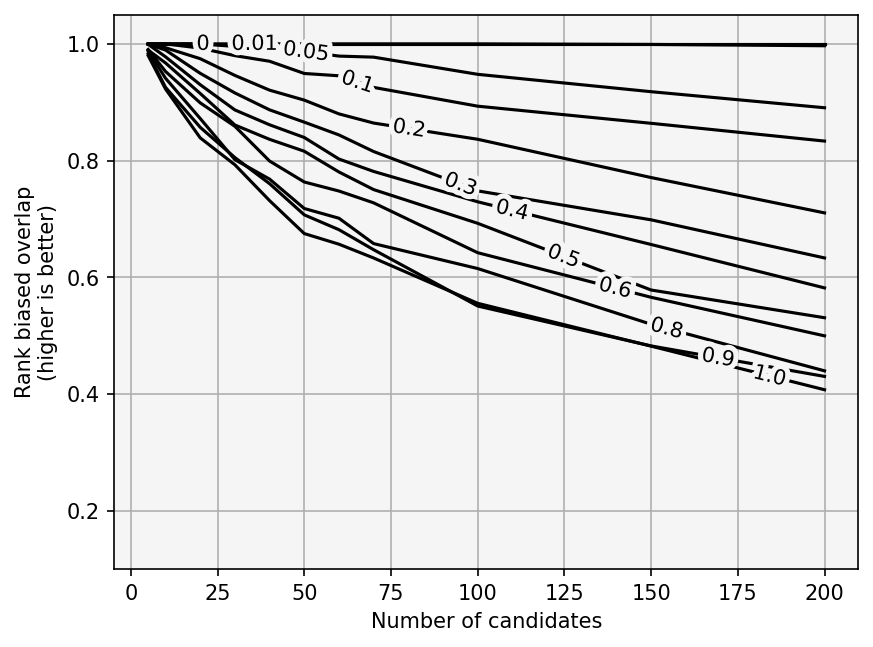}
            \caption{Performance of ranking based on expected rank}
            \label{fig:methods:synth:uncertain}
        \end{subfigure}
    \caption{Performance comparison in ranking using Rank Biased Overlap (RBO) metric for deterministic and probabilistic cases as a function of noise over 30 trials.}
    \label{fig:methods:synth}
    \Description{Image comparing performance of ranking between deterministic and probabilistic ranking methods.}
\end{figure*}

\subsection{User experience}

We develop a prototype of the system that computes the measures described above. A user experience to allow for exploration of matched candidates was designed. Figure \ref{fig:methods:ux:job} shows how recruiters can manage job requisitions through their lifecycle and then explore matched candidates (Figure \ref{fig:methods:ux:candidates}). As recruiters are generally familiar with ranked lists for candidates, our design follows this and uses a toggle to turn on humble AI principles. In this mode, the primary table is shown based on expected rank. As the expected rank differs considerably from the deterministic rank, a different set of candidates are surfaced. A second table that provides high entropy candidates is also shown. Users can change the proportion of high entropy candidates to see to increase (or decrease) exploration.

\subsection{A synthetic case}

The developments proposed in this paper so far implicitly state that point estimates of AI systems can lead to arbitrary rankings in hiring scenarios. Using humble AI principles, we have looked at post hoc uncertainty quantification and entropy to generate more robust rankings. To evaluate this, we conduct synthetic experiments. 

For a set of candidates for whom a true ranking is assumed to be known, their scoring function is perturbed with a known amount of noise. The scores generated are drawn from a normal distribution around the true value. We compare the rankings estimated from this noisy scoring function to the true ranking. One can expect that as the noise increase, the rankings diverge from the true ranking. This can be seen in Figure \ref{fig:methods:synth:deterministic}. The rank biased overlap (RBO) metric is used to compare the true and estimated rankings (higher is better). When one includes the distribution of ranks and uses average rank instead, the deviations from true rankings are considerably reduced as seen in Figure \ref{fig:methods:synth:uncertain}. The added effort of considering uncertain therefore has value in performance. To tie this experiment back to hiring, note that the scoring function $f_y$ used by AI systems for recruitment are indeed noisy in practice. Several factors, from non-traditional backgrounds, bias due to pedigree of institutions, manner in which expertise is mentioned all lead to noisy estimates of `true' fit.

\section{Discussion}
\label{sec:results}

\begin{figure*}[ht]
    \centering
    \captionsetup[subfigure]{justification=centering}
    \begin{subfigure}[t]{0.48\textwidth}
            \centering
            \includegraphics[height=70mm, keepaspectratio]{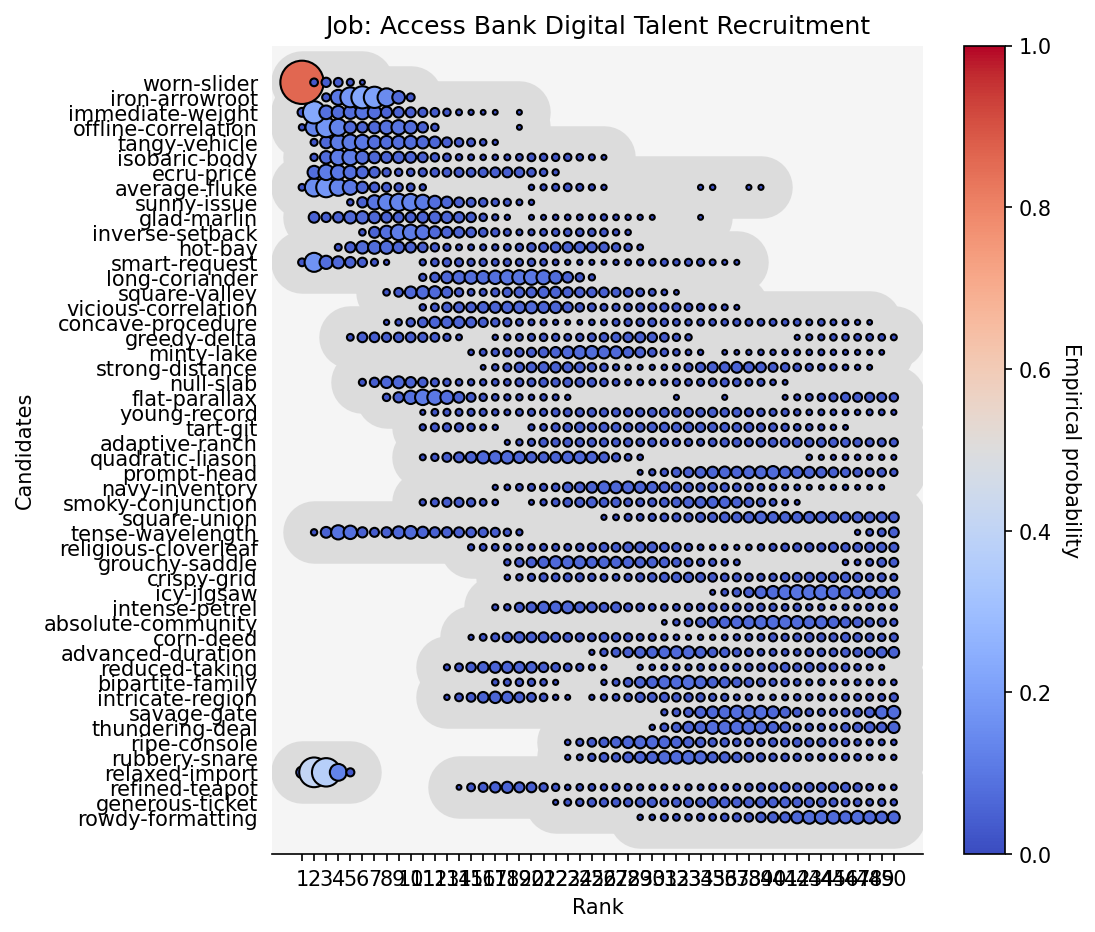}
        \end{subfigure}
    \begin{subfigure}[t]{0.48\textwidth}
            \centering
            \includegraphics[height=70mm, keepaspectratio]{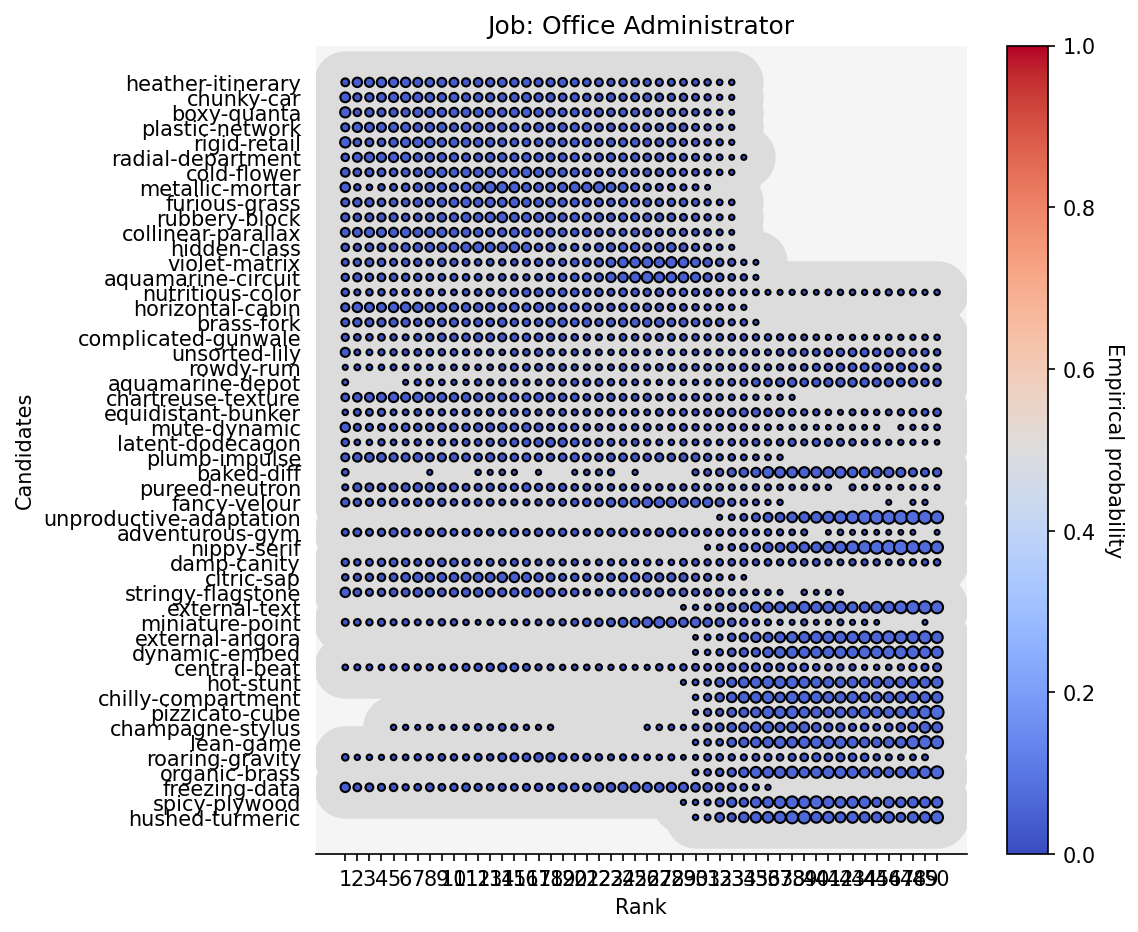}
        \end{subfigure}
    \caption{Selected rank sets for job titles (see Figures \ref{fig:app:ranksets} in appendix for more examples).}
    \label{fig:results:ranksets}
    \Description{Image showing rank sets for various job specifications.}
\end{figure*}

We run our experiments for several job specifications for multiple sectors and assume a candidate pool of 1000 candidates. We additionally assume that recruiters are shown 50 candidate profiles to make screening decisions, i.e. we are interested in determining the top-50 candidates from the pool. Figure \ref{fig:results:ranksets} shows selected examples. While point estimates from AI systems would have shown a strict order, accounting for uncertainty shows that most candidates are indistinguishable (judging by their rank overlap). Top performing candidates emerge as clusters. 

There is a very little correspondence between `deterministic' ranking derived from a single score from an AI system and from that seeking to capture the uncertainty in scoring (see Table \ref{tab:app:similarity} in the appendix for several measures comparing the two approaches). These results validate our central argument for caution in deployment of AI models for hiring. 

\paragraph{Limitations} Validation of uncertainties in this context is challenging. Unlike classical machine learning where one can compare prediction confidence and relate it to performance (generally a well calibrated model should have higher performance when it has high confidence), in the hiring scenario, the `true' outcome of a screening decision is not known. Indeed, the literature is sparse on the long term outcomes of hiring tools and their decisions. We offer some validation through our synthetic case where the true ranking is assumed to be known.

\section{User studies}

We briefly report on a focus group workshop held over half a day with 30 stakeholders including recruiters, HR professionals, and policy makers. The specialists describe the overall process from job requisitions, to requirements, to candidate screenings to be very iterative. The recruiters are often going back to the business to clarify/amend specifications. They also refine criteria as they speak to candidates. All decisions are typically made in time-poor settings, so recruiters have limited resources to conduct detailed reviews.

The specialists were receptive to main premise of humble AI principles. In our current designs, we separate out candidates with high mean rank and those with high uncertainties (Figure \ref{fig:methods:ux:candidates}). One participant pointed out that this may be misleading, as the uncertainty which we would like to attribute to the AI model can be attributed to the candidate. Revisions of this design could remove this distinction and surface candidates from both lists jointly. 

Specialists pointed out \emph{``reading between the lines''} when reviewing CVs, e.g. CVs with KPIs and focus on outcomes may indicate structural thinking that are more desirable. Automated parsers will likely miss this as a signal. Much of the discussion was focused on \emph{``soft skills''} and how hiring platforms, humble or otherwise, may being to consider this for automation. Specialists from the platform operator suggested that this information is only mentioned in the affirmative. Like \emph{``culture fit''} it is difficult to quantify and assess in an automated manner. Lastly, it was noted that the proposed humble AI tool is likely to warrant user training and careful design of the user experience to be effective. 

\begin{acks} 
This work was funded  by European Union’s Horizon Europe research and innovation programme under grant agreement no. 101070568 (AutoFair).
\end{acks}

\bibliographystyle{ACM-Reference-Format}
\bibliography{acmart}

\appendix
\section{Selected examples}
\begin{table*}[hbt!]
    \centering
\begin{tabular}{lrrr}
\toprule
Job Title & Similarity & RBO & Mean Entropy \\
\midrule
ReactJS Developer & 0.136 & 0.224 & 3.496 \\
Pharmaceutical Warehouse Associate & 0.695 & 0.714 & 2.889 \\
Payroll Implementation Specialist & 0.250 & 0.229 & 3.393 \\
Automotive Service Advisor & 0.190 & 0.147 & 3.507 \\
Senior Mobile Engineer (iOS) & 0.235 & 0.176 & 3.311 \\
Associate, Senior Associate, Policy Advisor, Manager & 0.163 & 0.102 & 3.467 \\
SALES EXECUTIVE & 0.010 & 0.002 & 3.177 \\
Business Development \& Sponsorships Manager & 0.000 & 0.000 & 3.089 \\
Expert SubtitlingTranslator: English to Galician, Basque, Catalan & 0.124 & 0.079 & 3.719 \\
General Engineering & 0.299 & 0.244 & 2.907 \\
Access Bank Digital Talent Recruitment & 0.053 & 0.025 & 3.039 \\
Office Administrator & 0.316 & 0.163 & 3.474 \\
Face to Face Language Trainer (m/f/d) & 0.042 & 0.020 & 3.171 \\
Clinical Content Builder & 0.124 & 0.194 & 2.673 \\
Data Engineer & 0.176 & 0.242 & 3.152 \\
Data Engineer & 0.042 & 0.051 & 3.214 \\
\bottomrule
\end{tabular}
    \caption{Comparing deterministic ranks to probabilistic rankings. Similarity is Jaccard Similarity. RBO - is Rank-biased overlap metric.}
    \label{tab:app:similarity}
\end{table*}

Table \ref{tab:app:similarity} compares the performance of rank sets relative to rankings determined by a single point estimate. Generally the agreement between the two is poor, indicating that the surfaced candidates can be arbitrary.

\begin{figure*}[hbt!]
    \centering
    \captionsetup[subfigure]{justification=centering}
    \begin{subfigure}[t]{0.45\textwidth}
            \centering
            \includegraphics[width=\linewidth,height=50mm, keepaspectratio]{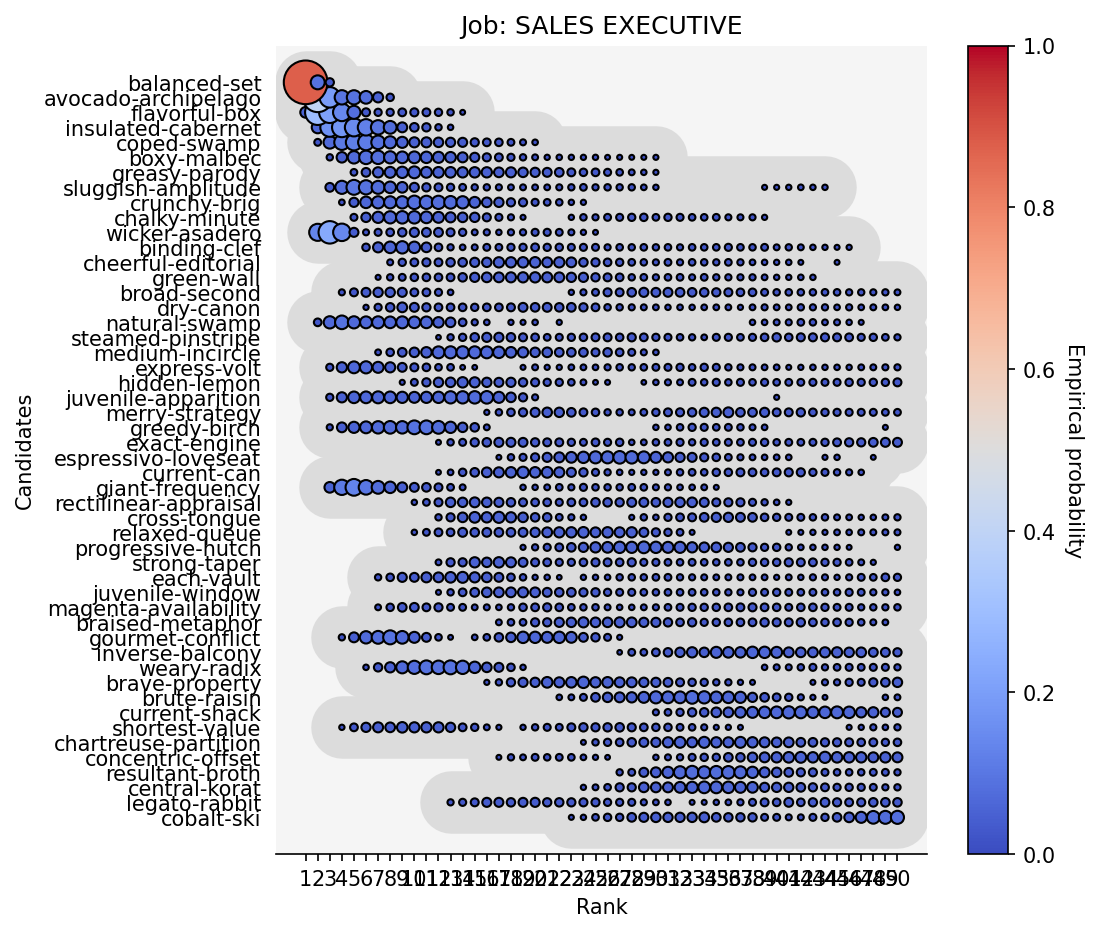}
        \end{subfigure}
    \begin{subfigure}[t]{0.45\textwidth}
            \centering
            \includegraphics[width=\linewidth,height=50mm, keepaspectratio]{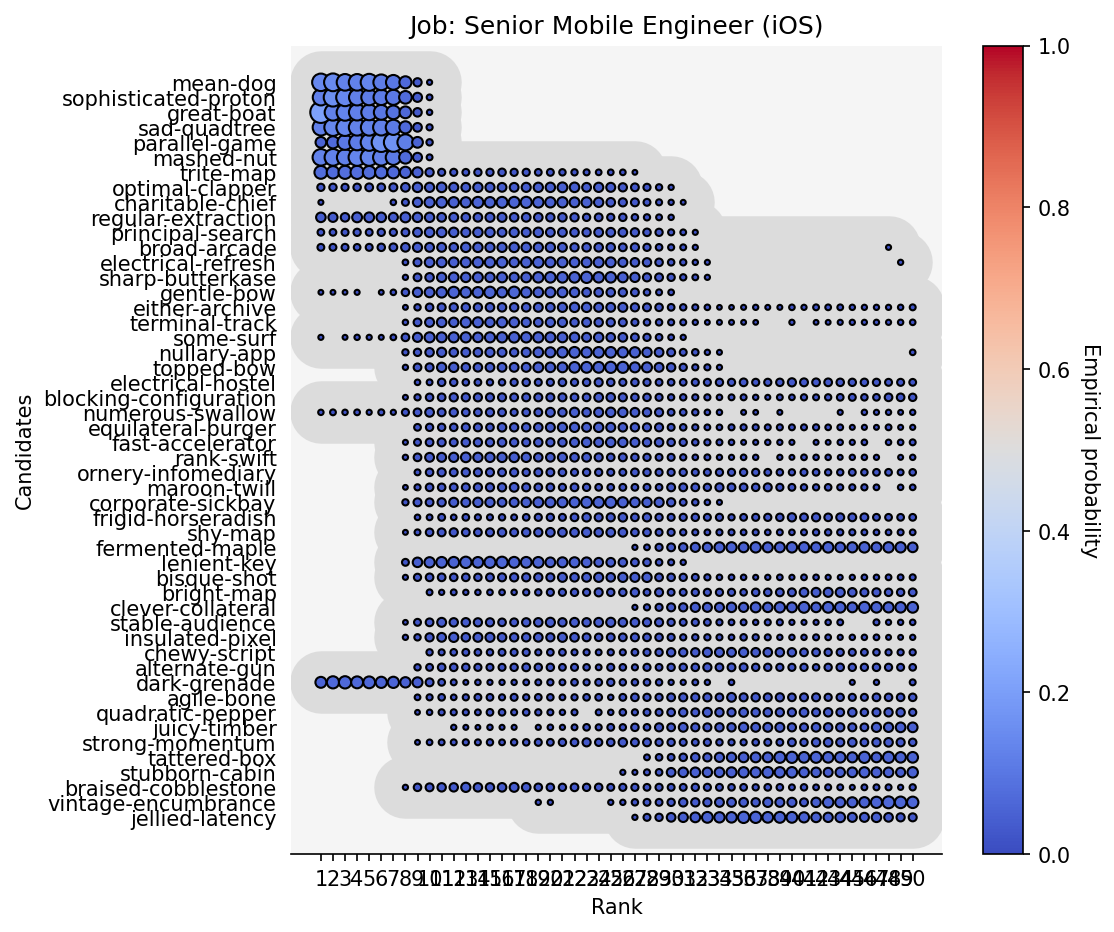}
        \end{subfigure} \\
            \begin{subfigure}[t]{0.45\textwidth}
            \centering
            \includegraphics[width=\linewidth,height=50mm, keepaspectratio]{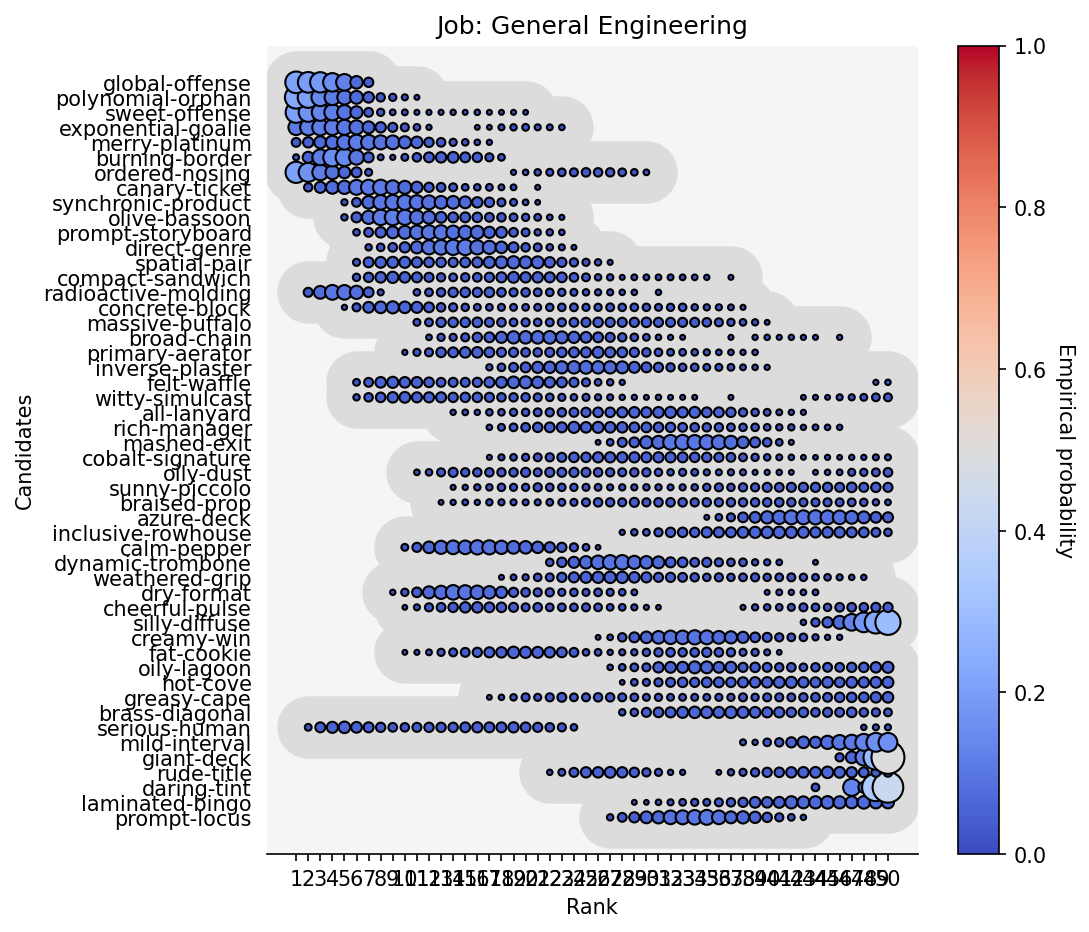}
        \end{subfigure}
    \begin{subfigure}[t]{0.45\textwidth}
            \centering
            \includegraphics[width=\linewidth,height=50mm, keepaspectratio]{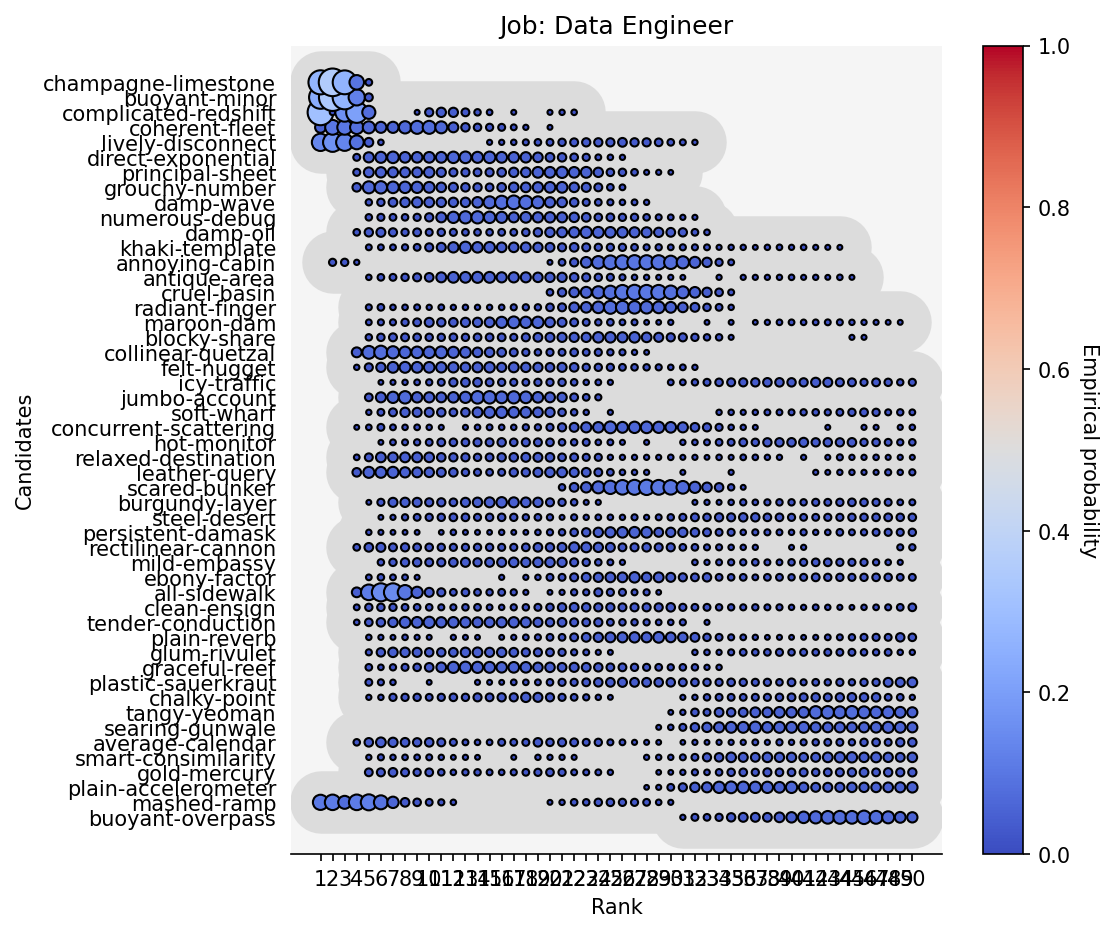}
        \end{subfigure} \\
        \begin{subfigure}[t]{0.45\textwidth}
            \centering
            \includegraphics[width=\linewidth,height=50mm, keepaspectratio]{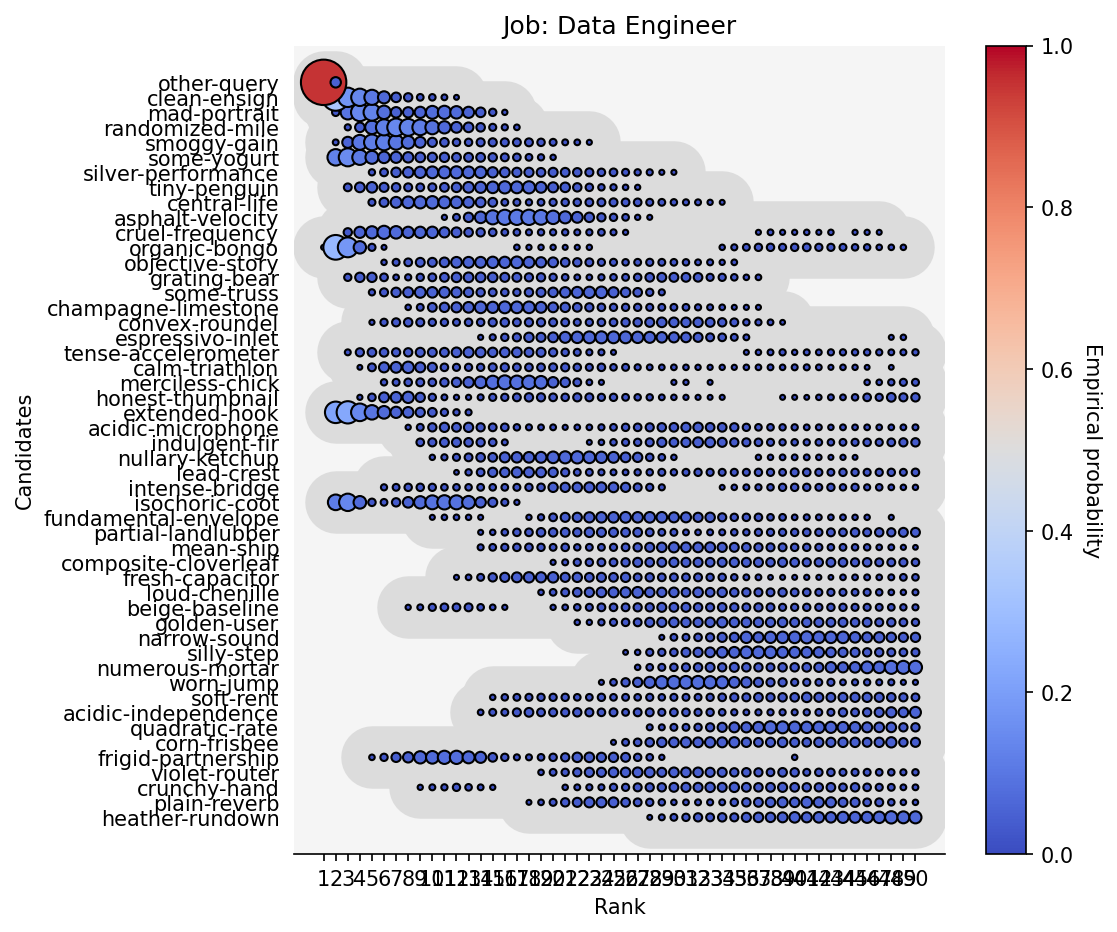}
        \end{subfigure}
    \begin{subfigure}[t]{0.45\textwidth}
            \centering
            \includegraphics[width=\linewidth,height=50mm, keepaspectratio]{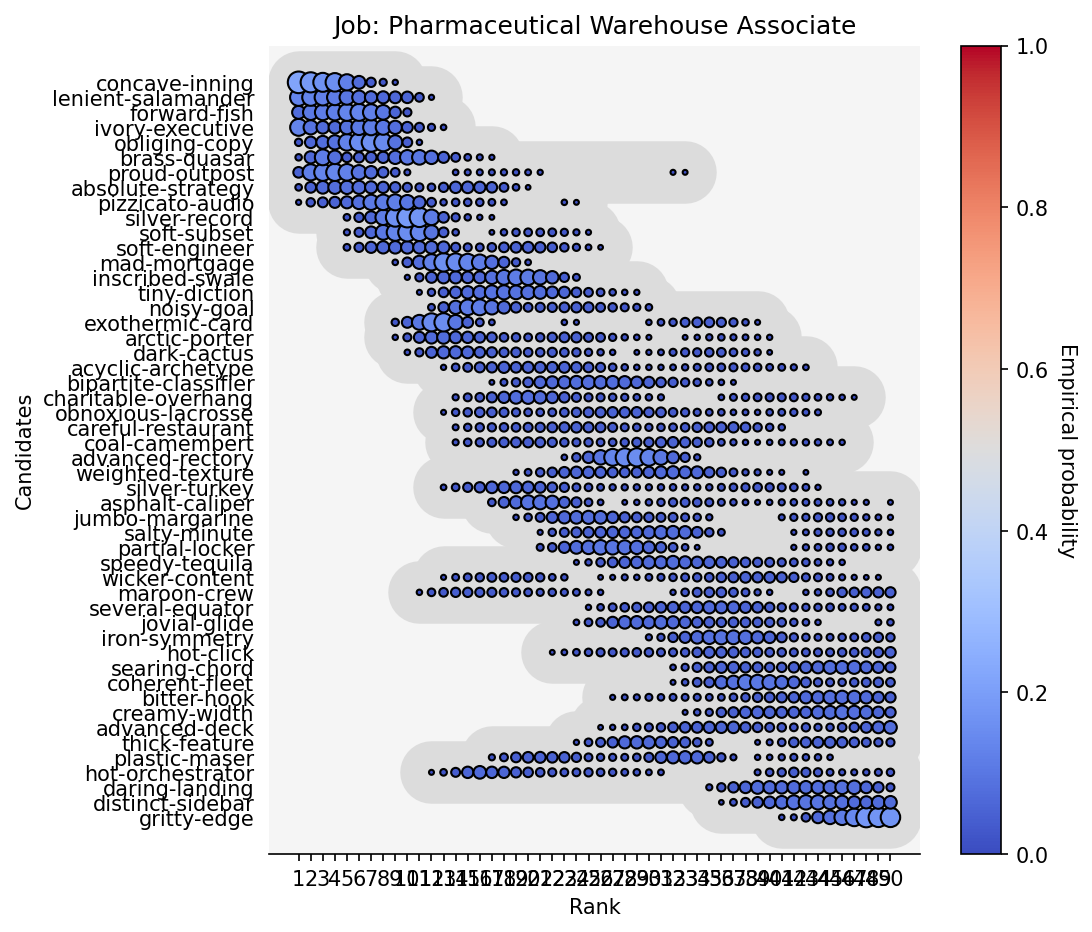}
        \end{subfigure} \\
        \begin{subfigure}[t]{0.45\textwidth}
            \centering
            \includegraphics[width=\linewidth,height=50mm, keepaspectratio]{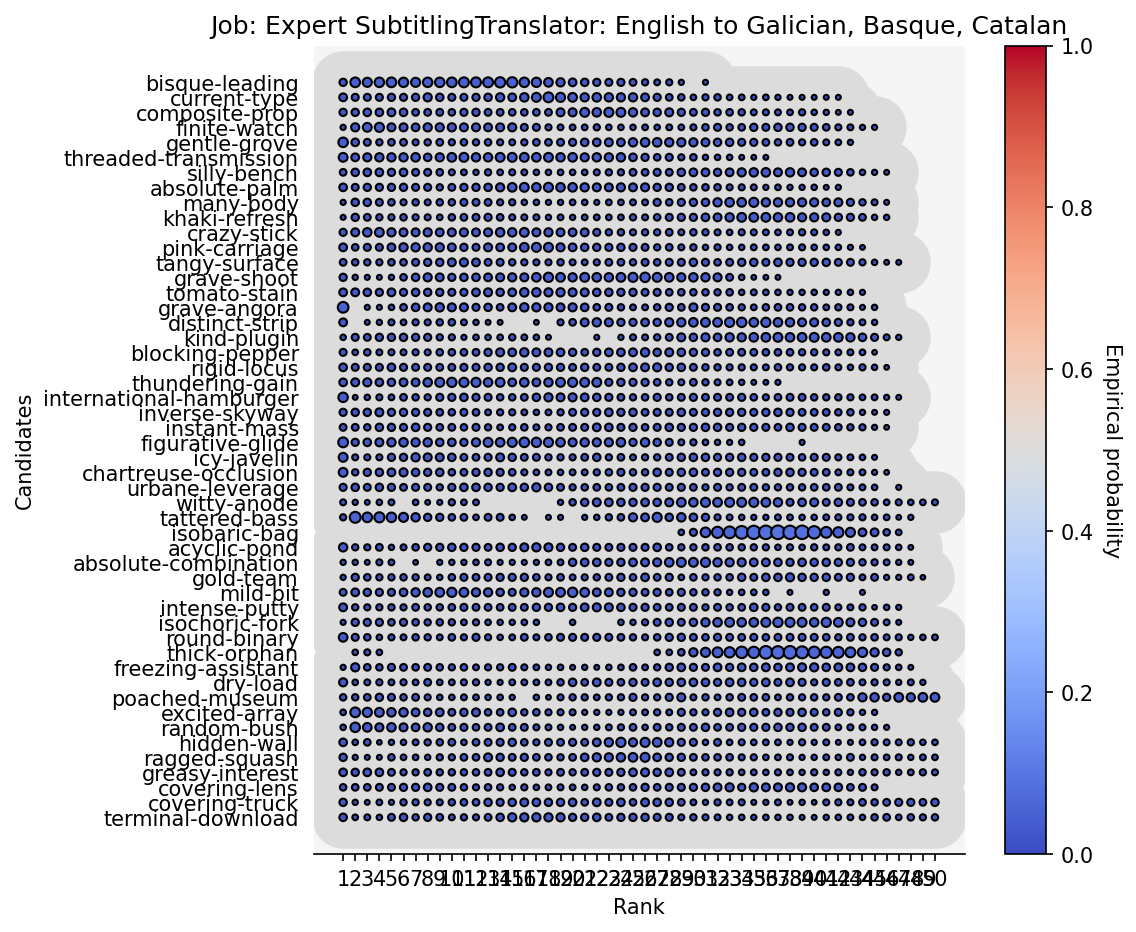}
        \end{subfigure}
    \begin{subfigure}[t]{0.45\textwidth}
            \centering
            \includegraphics[width=\linewidth,height=50mm, keepaspectratio]{figures/12.png}
        \end{subfigure} \\
    \caption{Selected rank sets for jobs}
    \label{fig:app:ranksets}
    \Description{Image showing rank sets for various job specifications.}
\end{figure*}

Figure \ref{fig:app:ranksets} shows additional examples of generated rank sets using our post hoc procedure. 

\section{User study details}

The user study was approved by Lancaster University's institutional review board. All participants of the focus group received a participant information sheet and signed a consent form agreeing to be recorded and content used for research purposes. The key ideas and a prototype system was presented to the group as part of a project living lab. Following a presentation, a group discussion took place which followed the semi-structured questionnaire with the following questions that were put to the audience. 
\begin{itemize}
    \item What recruitment challenges do these visualisations reveal? 
    \item How do you propose resolving these challenges? 
    \item What do you like about this approach?
    \begin{itemize}
        \item  What problems does it solve?
    \end{itemize}
    \item What do you dislike about this approach? 
    \begin{itemize}
        \item Does it create any new problems?
    \end{itemize}
    
\end{itemize}

\end{document}